\documentclass{article}

\usepackage[final,dblblindworkshop]{neurips_2026}
\workshoptitle{Agents in the Wild: Safety, Security, and Beyond}

\usepackage[utf8]{inputenc}
\usepackage[T1]{fontenc}
\usepackage{url}
\usepackage{booktabs}
\usepackage{tabularx}
\newcolumntype{Y}{>{\raggedright\arraybackslash}X}
\usepackage{amsfonts}
\usepackage{nicefrac}
\usepackage{microtype}
\usepackage{xcolor}
\usepackage{hyperref}
\usepackage{graphicx}
\usepackage{amsmath}
\usepackage{amssymb}
\usepackage{enumitem}
\usepackage{float}
\usepackage{placeins}
\usepackage{flafter} 
\usepackage{tikz}
\usetikzlibrary{positioning, arrows.meta, decorations.pathreplacing, calc}

\graphicspath{{figures/}}

\title{Prompted to Discriminate: Generalizing\\
Malicious-Input Probes in the Wild}

\author{%
  Elad David\\ Zenity\\ Tel Aviv, Israel\\ \texttt{eladd@zenity.io}%
  \And
  Max Fomin\\ Zenity\\ Tel Aviv, Israel\\ \texttt{maxf@zenity.io}%
}

\begin{document}
\maketitle

\begin{abstract}
LLM agents increasingly rely on activation probes as runtime monitors for prompt
injection, jailbreaks, and unsafe requests, reading the model's own hidden state to
catch a harmful input before the agent acts on it. A cheap and increasingly common move, borrowed from LLM-as-judge prompting, is to append a short classification instruction after the
user's turn and read the probe at that point, to \emph{try} to sharpen it: the instruction asks the model to represent the incoming request \emph{as a
class}, concentrating the signal the probe must separate, at negligible serving cost. But does the wording of that suffix matter, and does its benefit hold \emph{in the wild}, on
attack types the probe never saw in training? That out-of-distribution regime is the one a
deployed monitor faces. We test this with a controlled ladder of post-user suffixes under strict
leave-one-dataset-out (LODO) evaluation across 13 safety benchmarks (jailbreak, injection, and benign chat) and three open-weight model families
(Llama-3.1-8B, Qwen3.5-9B, Gemma-4-12B). On a single-position probe, a classification suffix \emph{consistently improves}
out-of-distribution detection over the no-suffix baseline (up to ${\sim}4$ AUC points); yet
\emph{which} suffix matters: prompting the model to \emph{classify} the input, even into
content-free labels, reliably wins, whereas an off-topic or merely-attentive suffix helps
little. The gain comes from the classification \emph{format}, not the named criterion: a
content-free suffix does as well as the real malicious/benign one, with the criterion adding
precision only at strict thresholds. This is not an artifact of the single-position read: the
benefit carries to the multi-position pooling probes used in production (attention, multi-max,
MLP), though the best-performing suffix there is readout-dependent. Served through a KV-cache fork,
it is a cheap drop-in for any activation-probe monitor, though not an automatic win: \emph{which}
suffix helps, and by how much, depends on the model and the readout.
\end{abstract}

\section{Introduction}
\label{sec:intro}
LLM agents increasingly rely on activation-based probes as runtime monitors for
prompt injection, unsafe intent, and policy violations
\citep{ZouRepE2023,TaskTracker,PIShield}: a probe on the model's hidden state flags a
malicious request before the agent acts, riding the model's own forward pass rather than a
separate classifier; recent production monitors pool over many positions
\citep{KramarGeminiProbes2026}. A central
question for deploying such a monitor is how to make it read well.
\citet{TillmanMossing2025} show that appending a task-description instruction
(\emph{prompted probing}) improves out-of-distribution safety monitoring, but Tillman \& Mossing's
instructions are semantically meaningful throughout and the mechanism is left
open. We ask the question that determines how the monitor should be
built: \emph{what part of the instruction does the work: the fact that it induces
a classification, or the specific criterion it names?}

We answer it with a \textbf{designed ladder of post-user suffixes}, each isolating one
ingredient (Table~\ref{tab:ladder-design}). We study \emph{post-user} suffixes specifically because they deploy for
free: appended after the user turn on a KV-cache fork (adopted for the simplicity and
perturbation-robustness \citet{LatentUndertow2026} report), read once, then reverted
(\S\ref{sec:suffix}), so the monitor adds no separate forward pass and inherits that
vehicle's position stability and robustness to input perturbations.

We make three contributions:
\begin{enumerate}[noitemsep,topsep=2pt]
  \item \textbf{We extend prompted probing into the wild, and decompose it}
    (\S\ref{sec:ladder}). \citet{TillmanMossing2025} showed a task instruction improves
    out-of-distribution monitoring on one closed model; we carry that analysis to three
    open-weight families under strict leave-one-dataset-out, and go further with a length-matched
    suffix \emph{ladder} that isolates which ingredient of the instruction does the work. The
    classification suffix consistently improves detection (up to ${\sim}4$ AUC points; positive on
    $22$ of $27$ model--dataset cells, App.~\ref{app:perds}).
  \item \textbf{The classification \emph{format}, not the named criterion, carries the ranking
    gain} (\S\ref{sec:ladder}). Being \emph{asked to classify} is what transfers: a content-free
    A/B suffix is statistically equivalent to the intent suffix (${<}1$ AUC point apart; TOST
    $p{\le}0.005$; Llama and Gemma), while off-topic or attend-only suffixes do no better than
    no suffix. The named criterion adds precision only at strict thresholds.
  \item \textbf{The benefit holds across readout architectures} (\S\ref{sec:arch}). It survives
    the move from a single-position read to the multi-position pooling probes (attention,
    multi-max, MLP) used at scale ($92\%$ of readout$\times$suffix cells beat no-suffix), so the
    gain is a property of the representation the suffix induces, not of one probe head.
    The best-performing suffix is readout-dependent, making suffix choice a per-probe design knob.
\end{enumerate}

\begin{table}[H]
  \centering
  \caption{\textbf{The suffix ladder (experimental design).} Each suffix adds one
    ingredient; comparing adjacent suffixes isolates it.}
  \label{tab:ladder-design}
  \small
  \begin{tabularx}{\linewidth}{@{}l>{\raggedright\arraybackslash}p{0.27\linewidth}Y@{}}
    \toprule
    Suffix & Adds & Isolates \\
    \midrule
    \texttt{none}              & ---                      & baseline \\
    \texttt{irrelevant}        & off-topic text           & is \emph{any} appended text enough? \\
    \texttt{reflect}           & instruction to attend    & is attending (no decision) enough? \\
    \texttt{decide}            & a binary decision (prose)& does deciding help without a template? \\
    \texttt{classify\_grammar/length} & + output template, unrelated criterion & does the criterion matter? \\
    \texttt{A/B}               & content-free labels      & does criterion \emph{content} matter? \\
    \texttt{intent}            & malicious/benign criterion & the ``meaningful'' criterion \\
    \bottomrule
  \end{tabularx}
\end{table}

\paragraph{Scope.} Building on prompted probing \citep{TillmanMossing2025} and the
KV-fork suffix \citep{LatentUndertow2026}, our contribution is to establish, by controlled
ablation, \emph{what in the suffix} drives the gain and how broadly it holds: across
suffixes, models, and probe architectures. Table~\ref{tab:vs_tm} summarizes what we add
over Tillman \& Mossing.

\begin{table}[H]
  \centering
  \caption{\textbf{Delta over \citet{TillmanMossing2025}.} Every row is backed in
    this paper. (T\&M report AUROC on a single closed model, GPT-4o, with limited
    within-task distribution shift on one dataset.)}
  \label{tab:vs_tm}
  \small
  \begin{tabularx}{\linewidth}{@{}>{\raggedright\arraybackslash}p{0.2\linewidth}YY@{}}
    \toprule
    Axis & Tillman \& Mossing 2025 & \textbf{This work} \\
    \midrule
    What in the prompt matters & meaningful instruction (not decomposed) & format-vs-content ladder \\
    Generalization test        & limited within-task shift, 1 dataset    & cross-dataset LODO, 13 benchmarks \\
    Models                     & GPT-4o (one, closed)                    & 3 open families (Llama/Qwen/Gemma) \\
    Robustness                 & single layer/position                   & across layers \emph{and} readout architectures (single-position to pooling) \\
    Content sensitivity        & ---                                     & content-tailoring overfits OOD \\
    Deployment                 & prompt-side instruction                 & KV-fork suffix: single forward pass, no re-prefill \\
    \bottomrule
  \end{tabularx}
\end{table}

Concretely, every suffix shares the stem \texttt{CRITICAL REMINDER --} and differs
only in what follows. Four verbatim examples (the full set is in
App.~\ref{app:perds}):
\begin{quote}\small\ttfamily
\textbf{intent:} …start your response with `Input is $\langle$class$\rangle$' where
classes are malicious/benign.\\
\textbf{A/B:} …classes are A/B.\\
\textbf{reflect:} …take a moment to read the message above carefully and answer
completely.\\
\textbf{irrelevant:} …here is an unrelated fact: the ocean covers most of the planet.
\end{quote}

\section{Background and Related Work}
\label{sec:related}
\paragraph{Prompted probing.}
\citet{TillmanMossing2025} introduce instruction-conditioned probing and show it
improves OOD safety monitoring on a GPT-4o snapshot (AUROC; linear vs.\ prompted
vs.\ SAE probing). Activation probing has progressed from reading a concept as a linear direction in the
hidden state \citep{ZouRepE2023} (as single-position injection detectors still do,
\citep{TaskTracker,PIShield}) to the multi-position \emph{pooling} probes (attention,
multi-max, MLP) that \citet{KramarGeminiProbes2026} deploy at production scale. That line
develops the probe \emph{architecture}; prompted probing is orthogonal, acting on the
input. We combine the two, taking prompted probing across these SOTA readout
architectures, from a single position to the advanced multi-position probes.

\paragraph{Do these probes read the concept, or surface features?}
\citet{Wang2025FalseSecurity} show malicious-input probes can reach $>$98\% in-distribution yet
learn trigger words and instructional patterns and fail to generalize. This bears on us two
ways. Our content-free \texttt{A/B} control answers the trigger-word half: the gain survives when
the criterion carries no attack vocabulary, so it is not that shortcut. On the
instructional-pattern half, our suffix \emph{is} such a pattern, so we make no claim that the probe reads
ground-truth harmfulness, only that a classification \emph{format} yields a more transferable
readout on our confound-controlled benchmarks.

\paragraph{Format over content.}
\citet{Min2022Rethinking} show in-context learning is largely insensitive to label
\emph{content} (format over content), behaviorally, via demonstrations; prompts are
broadly format-sensitive \citep{SclarFormatSpread2024,MizrahiMultiPrompt2024}. Our
result is the activation-level analogue for a safety probe: prompting the model to
\emph{classify} the input (even into content-free labels) reshapes its representation
to be more separable across datasets, largely independent of the criterion named. We state
this for the models with a gain to carry (Llama, Gemma); on Qwen it is small (near-ceiling).

\paragraph{Cross-dataset evaluation.}
\citet{Fomin2026BenchmarksLie} show malicious classifiers collapse under LODO
relative to in-distribution CV (pooled AUC gap $8$--$16$ points) on overlapping
model families, but explicitly find that SAE-feature dataset shortcuts do
\emph{not} causally account for the gap and never vary how the classifier is
prompted. We adopt their LODO rigor but ask a different question: how the readout is prompted,
not which trained-probe features it exploits. Our aim is a better-ranking probe, not a
claim to eliminate the in-distribution gap they measure.

\paragraph{The deployment driver.}
\citet{LatentUndertow2026} introduce the KV-fork post-user suffix (a stable single-position
readout at near-zero cost) and frame agent monitoring around the low-FPR operating point
rather than aggregate AUROC. They show a \emph{generic} suffix makes the readout robust to input
perturbations, but do not ask what a \emph{classification} instruction adds. We sit between the
two: we take the prompted-probing question of \citet{TillmanMossing2025} and the suffix vehicle
and operating-point lens of \citet{LatentUndertow2026}, and resolve the instruction into a
controlled ladder separating format from content under LODO.

\section{Setup}
\label{sec:setup}
\paragraph{Models and probes.} Llama-3.1-8B, Qwen3.5-9B, and Gemma-4-12B, read at late layers
(${\sim}2/3$ of depth, where the malicious/benign distinction is most linearly accessible; exact
layers in App.~\ref{app:probes}). Our default readout is a frozen single-position linear probe (L2-logistic,
standardized, $C{=}0.1$) at the last suffix token; \S\ref{sec:arch} additionally evaluates
the advanced multi-position \emph{pooling} probes (attention, multi-max, MLP). We check
robustness across two layers per model (\S\ref{sec:measurement}); full probe and layer
settings in App.~\ref{app:probes}.
\paragraph{Suffix ladder.} Table~\ref{tab:ladder-design}; all suffixes share the
\texttt{CRITICAL REMINDER -- } stem and are length-controlled, so adjacent
comparisons isolate a single ingredient rather than length or mere presence of
text. The main suffix (\texttt{intent}, or its content-free twin \texttt{A/B}) is 29 tokens.
\paragraph{Benchmarks.} 13 chat safety, jailbreak, and injection benchmarks (full list in App.~\ref{app:data}), evaluated LODO: each dataset's malicious examples are scored
against one shared, balanced benign pool. \textbf{$N{=}2000$ prompts per dataset} (fewer where a dataset is smaller).

\subsection{The suffix and its deployment vehicle}
\label{sec:suffix}
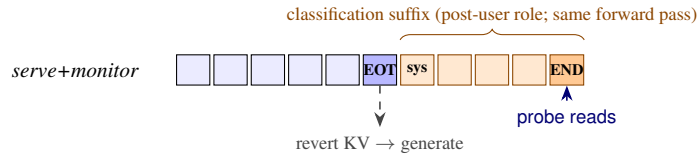
\begin{figure}[H]
  \centering
  \begin{tikzpicture}[
    tok/.style={draw, minimum width=4.4mm, minimum height=4mm, inner sep=0pt,
                font=\tiny, line width=0.4pt},
    user/.style={tok, fill=blue!8},
    eot/.style={tok, fill=blue!28, font=\tiny\bfseries},
    suffix/.style={tok, fill=orange!18, draw=orange!55!black},
    suff-end/.style={suffix, fill=orange!42, font=\tiny\bfseries},
    arr/.style={-Stealth, semithick},
    rev/.style={-Stealth, semithick, dashed, color=gray!55!black},
    rd/.style={-Stealth, semithick, color=blue!45!black},
    rd-lbl/.style={font=\scriptsize\sffamily, color=blue!45!black},
    row-lbl/.style={font=\footnotesize\itshape, anchor=east},
  ]
  \node[row-lbl] at (-0.1, 0) {serve+monitor};
  \node[user] at (0.5, 0) (u1) {};
  \node[user, right=0.4mm of u1] (u2) {};
  \node[user, right=0.4mm of u2] (u3) {};
  \node[user, right=0.4mm of u3] (u4) {};
  \node[user, right=0.4mm of u4] (u5) {};
  \node[eot, right=0.4mm of u5] (eot) {EOT};
  \node[suffix, right=0.4mm of eot, font=\tiny\bfseries] (srole) {sys};
  \node[suffix, right=0.4mm of srole] (s1) {};
  \node[suffix, right=0.4mm of s1] (s2) {};
  \node[suffix, right=0.4mm of s2] (s3) {};
  \node[suff-end, right=0.4mm of s3] (send) {END};
  \node[rd-lbl, below=2mm of send] (rd) {probe reads};
  \draw[rd] (rd.north) -- (send.south);
  \draw[rev] (eot.south) -- ++(0, -0.55)
    node[below, font=\scriptsize, color=gray!55!black] {revert KV $\to$ generate};
  \draw[decorate, decoration={brace, amplitude=1.4mm, raise=1.4mm},
        color=orange!55!black]
    ($(srole.north west)+(0,0.5mm)$) -- ($(send.north east)+(0,0.5mm)$)
    node[midway, above=2.2mm, font=\scriptsize, color=orange!55!black]
    {classification suffix (post-user role; same forward pass)};
  \end{tikzpicture}
  \caption{\textbf{The classification suffix on the KV-cache fork.} A short classification
    instruction is appended after the user turn; the probe reads at end-of-suffix; the cache is
    reverted so generation proceeds unchanged. We adopt the KV-fork vehicle from
    \citet{LatentUndertow2026} and study what the suffix should \emph{say}.}
  \label{fig:kvfork}
\end{figure}

The suffix rides the post-user KV-cache fork of Figure~\ref{fig:kvfork}: a short
classification instruction is appended after the user turn, the probe reads at
end-of-suffix, and the cache is reverted so the served response is byte-identical to
no-suffix serving: monitoring rides the model's existing forward pass, not a second one.

We \emph{inherit} this vehicle's benefits rather than re-establish them: negligible
overhead (a 29-token suffix and a few MB of KV cache, with no second forward pass),
perturbation-robustness, and a readout-stabilization gain for a \emph{generic} suffix
\citep{LatentUndertow2026}, and ask the orthogonal question: what the suffix should
\emph{say}. Adopting it is a one-line change to an existing probe pipeline (same
architecture, same training data, one appended suffix) that keeps serving and monitoring in a
single forward pass, unlike prompt-reformatting \citep{TillmanMossing2025}.

\paragraph{Metrics.} Primary = \textbf{shared-benign AUC}: each malicious-containing
dataset's malicious examples are ranked against one fixed, shared benign pool. This is threshold-free and defined for all 9 malicious-containing datasets. Plus
\textbf{TPR@1\% FPR}, the strict operating point a deployed guardrail runs at
\citep{LatentUndertow2026,McKenzie2025HighStakes}. We tabulate deltas versus \texttt{none} (no suffix), but the
deployment-relevant comparison is versus a \emph{neutral} suffix (the relocation
a monitor may already run; \S\ref{sec:ladder}). All numbers are LODO out-of-fold, $N{=}2000$/dataset;
we report \textbf{per-dataset}; any aggregate is an equal-weight mean across datasets.

\section{Results and Analysis}
\label{sec:ladder}
We first analyze the single-position probe (\S\ref{sec:single}), then show the benefit extends
to the multi-position pooling probes used in production (\S\ref{sec:multipos}).

\subsection{Single-position probe}
\label{sec:single}
\begin{figure}[t]
  \centering
  \includegraphics[width=0.88\linewidth]{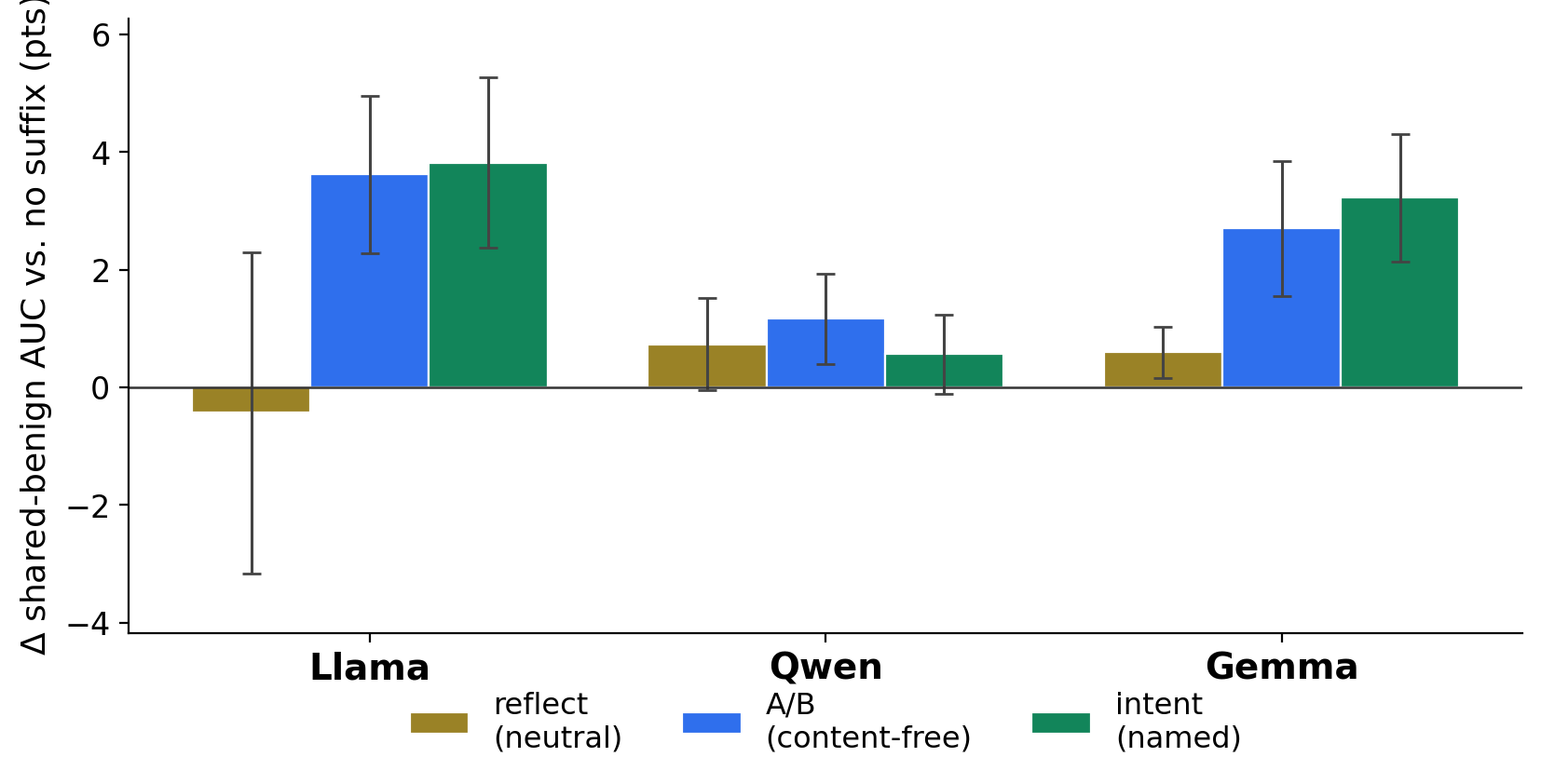}
  \caption{\textbf{A classification suffix lifts OOD detection; format, not criterion, carries
    the ranking gain.} $\Delta$ shared-benign AUC vs.\ no suffix (mean over 9 LODO datasets,
    $\pm$SEM). A neutral suffix (\texttt{reflect}) barely moves ranking; asking the model to
    \emph{classify} lifts it (Llama, Gemma), and content-free \texttt{A/B} matches named
    \texttt{intent}. Qwen is near-ceiling. Precision (the criterion's low-FPR gain): App.~\ref{app:tpr}.}
  \label{fig:ladder}
\end{figure}

A classification suffix consistently improves out-of-distribution detection: up to ${\sim}4$
shared-benign AUC points over no suffix, positive on 22 of 27 (model, dataset) cells
(Figure~\ref{fig:ladder}). \emph{What in it} does the work splits into two ingredients: the
classification \emph{format} carries the cross-dataset \emph{ranking} gain (shared-benign AUC),
while naming the real \emph{criterion} adds \emph{precision} at a strict false-positive rate.
The two dissociate cleanly: a content-free format matches the full suffix on ranking yet
leaves the precision gain on the table (App.~\ref{app:dissoc}). Throughout, we compare each
classification suffix both to \emph{no suffix} and to a \emph{neutral} suffix (\texttt{reflect}):
since appending any generic suffix already relocates the readout, the neutral suffix is the
stricter baseline for what \emph{classification} adds.

The ranking gain (Table~\ref{tab:ladder}) differs by model in magnitude: large on Llama and
Gemma, near-ceiling on Qwen. Naming the real criterion additionally sharpens recall at a strict
$1\%$-FPR operating point (App.~\ref{app:tpr}), the regime a deployed monitor runs at
\citep{LatentUndertow2026}; our recall there is in range of concurrent probe-monitoring under
shift (\citealp{McKenzie2025HighStakes}: $43\%$ at $1\%$ FPR vs.\ ${\sim}55\%$ for their best
LLM monitor).

\begin{table}[t]
  \centering
  \caption{\textbf{Suffix ladder: $\Delta$ shared-benign AUC vs.\ \texttt{none}} (points; LODO,
    $N{=}2000$; 9 malicious-containing datasets; paired Wilcoxon $^{*}p{<}0.05$). Content-free
    \texttt{A/B} $\approx$ named \texttt{intent} across models; on Llama the null suffixes sit at
    baseline, while Gemma responds to \emph{any} post-user suffix (relocation, \S\ref{sec:ladder}).
    Qwen is near-ceiling.}
  \label{tab:ladder}
  \small
  \begin{tabular}{lccc}
    \toprule
    Suffix & Llama & Qwen & Gemma \\
    \midrule
    \texttt{irrelevant} (append only)   & $-0.14$ & $+0.07$ & $\mathbf{+1.29^{*}}$ \\
    \texttt{reflect} (attend)           & $-0.44$ & $+0.76$ & $+0.61$ \\
    \texttt{decide} (prose decision)    & $+2.14$ & $-0.25$ & $+0.41$ \\
    \texttt{classify\_grammar}          & $+2.00$ & $+0.85$ & $+2.33$ \\
    \texttt{classify\_length}           & $-1.56$ & $+1.22$ & $\mathbf{+3.09^{*}}$ \\
    \texttt{A/B} (content-free)         & $\mathbf{+3.61^{*}}$ & $+1.18$ & $\mathbf{+2.71^{*}}$ \\
    \texttt{intent} (mal/benign)        & $\mathbf{+3.82^{*}}$ & $+0.59$ & $\mathbf{+3.23^{*}}$ \\
    \bottomrule
  \end{tabular}
\end{table}

\paragraph{Classification, not the criterion, carries the ranking gain.} Two comparisons
isolate the format. First, suffixes that add \emph{no} classification (off-topic text, or an
instruction merely to attend) leave the probe unmoved (Llama \texttt{irrelevant} $-0.14$,
\texttt{reflect} $-0.44$; n.s.), whereas asking the model to classify lifts it. Second, once it
classifies, the \emph{criterion} does not matter: a content-free \texttt{A/B} suffix is
statistically equivalent to the real \texttt{intent} one on shared-benign AUC (TOST, $\pm2$-pt
bound; Llama $p{=}0.004$, Gemma $p{=}0.005$). So the classification \emph{format}, not the named
criterion, produces the transferable ranking. (The criterion is not \emph{entirely} inert: two
arbitrary criteria disagree on Llama, \texttt{classify\_grammar} $+2.0$ vs.\
\texttt{classify\_length} $-1.6$; but a meaningful one adds nothing over a content-free one.)
Naming the real classes does buy low-FPR precision on top, at a single noisy operating point
(App.~\ref{app:tpr}).

\paragraph{The right baseline is a neutral suffix, not the raw probe.} A neutral post-user suffix
is essentially the KV-fork relocation \citet{LatentUndertow2026} already credit for robustness
(we use \texttt{reflect}); it, not the raw probe, is the baseline for ``what does a
\emph{classification} suffix add.'' Over it the ranking increment is modest but consistent: the
classification suffix wins $73\%$ of per-dataset comparisons (38/52; sign-test
$p{=}6{\times}10^{-4}$; Llama median $+1.8$, the larger mean inflated by \texttt{mosscap}); the
clean win over neutral is at strict precision (App.~\ref{app:tpr}).

\begin{figure}[t]
  \centering
  \includegraphics[width=0.62\linewidth]{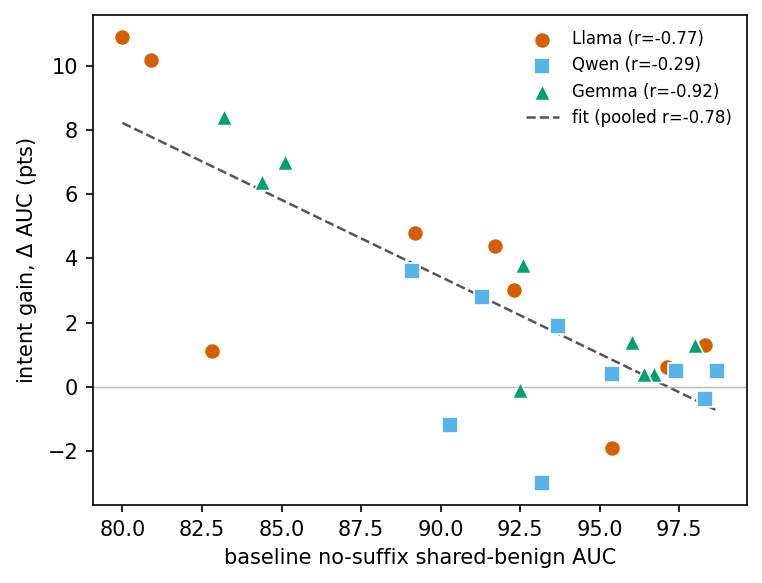}
  \caption{\textbf{The gain concentrates where the probe has headroom.} Each point is one
    (model, dataset): no-suffix shared-benign AUC ($x$) vs.\ the intent gain ($y$). Pooled
    $r{=}{-}0.78$ (Llama $-0.77$, Gemma $-0.92$, Qwen $-0.29$); part of the slope is a ceiling
    effect (bounded AUC), so we read it as a headroom regularity, not a law.}
  \label{fig:headroom}
\end{figure}

\paragraph{Magnitude tracks headroom, with a ceiling caveat.} The improvement is consistent: 22 of 27 (model, dataset) cells (sign-test $p{<}10^{-3}$), strong on Llama and Gemma, smaller on
Qwen, and concentrates where the base probe is weakest: the intent gain falls as the no-suffix
baseline rises (pooled $r{=}{-}0.78$, $p{=}1.4{\times}10^{-6}$; Fig.~\ref{fig:headroom}). Part of
this slope is mechanical (the gain is bounded above), and \emph{within} Qwen the trend is weak
($r{=}{-}0.29$), so headroom is only a partial account of Qwen's smaller effect; its readout
regime (thinking mode) is a confound we do not disentangle.

The gain lands on non-saturated datasets (Llama \texttt{deepset} $80\to90$, \texttt{harmbench}
$81\to91$) and not on near-ceiling ones, and the mean can mislead: \texttt{reflect}'s aggregate
``null'' is a cancellation that helps \texttt{deepset}/\texttt{harmbench} but crashes
\texttt{mosscap} ($83\to63$), so we report per-dataset throughout (App.~\ref{app:perds}).

Where the probe has headroom, we can characterize \emph{why} the suffix helps: it aligns an
otherwise-scattered per-source malicious direction into a shared axis (on Gemma, leave-one-source-out
transfer AUC $0.74\to0.93$; the classes visibly separate in a t-SNE of the readout), whereas on
models whose malicious axis is already formed without a suffix (Llama, Qwen; transfer $0.93$, $0.96$)
there is little to add, a representational reading of the headroom-gating seen throughout, detailed
in App.~\ref{app:mech}.

\subsection{Multi-position (pooling) probes}
\label{sec:multipos}\label{sec:arch}
Multi-position monitors increasingly read the hidden state not at a single position but by
\emph{pooling} over many (attention, multi-max, or an MLP head)
\citep{KramarGeminiProbes2026}. If the suffix's gain were an artifact of the
single-position readout we (and \citealp{TillmanMossing2025}) use, it need not survive
that shift. It does. Table~\ref{tab:arch} reports, for each readout, the mean suffix gain over no-suffix across the
five-rung ladder (best single rung in parentheses), holding the probe's training data and the
LODO protocol fixed: a suffix lifts \emph{every} readout we test, from the single-position
read to the advanced multi-position probes.

\begin{table}[t]
  \centering
  \caption{\textbf{The suffix helps across readout architectures.} Suffix gain over no-suffix in
    shared-benign LODO AUC (pp), primary layers, per pooling readout
    \citep{KramarGeminiProbes2026}. Each cell: mean over the five-rung ladder (best rung in
    parentheses). Averaged over the ladder, \textbf{22/24} model$\times$readout configs and
    \textbf{110/120} (readout$\times$rung) cells beat no-suffix (92\%), needing no rung selection.
    Full grid: Table~\ref{app:fullgrid}.}
  \label{tab:arch}
  \small
  \begin{tabular}{lccc}
    \toprule
    Readout & Gemma-4 & Qwen3.5 & Llama-3.1 \\
            & (L47)   & (L31)   & (L31) \\
    \midrule
    attention & $+1.7\ (+2.0)$ & $+0.8\ (+1.3)$ & $+2.3\ (+2.8)$ \\
    multi-max & $+4.5\ (+4.8)$ & $+0.4\ (+0.7)$ & $+2.0\ (+2.9)$ \\
    MLP       & $+1.5\ (+1.9)$ & $+0.9\ (+1.0)$ & $+1.6\ (+1.8)$ \\
    mean      & $+2.1\ (+2.4)$ & $+0.6\ (+0.7)$ & $+1.9\ (+2.2)$ \\
    \bottomrule
  \end{tabular}
\end{table}

The lift is model-modulated, largest where the single-position probe has headroom
(Gemma, Llama), smallest where it is already saturated (Qwen, baselines ${\sim}95$; its one
negative cell, in the full grid, is the most-saturated attention readout at its second
layer). Two things follow. First, the benefit is a property of the \emph{representation},
not of one probe head: any reasonable readout can exploit it. The format/criterion split holds
here too, not just at the single position: a content-free \texttt{A/B} suffix matches the named
\texttt{intent} one within $0.3$ AUC points on average across pooling readouts (23 of 24 cells
within $1$ pt; Table~\ref{app:fullgrid}). Second, and more practically,
\textbf{the best-performing rung is readout-dependent} (\texttt{length} and \texttt{reflect}
top more columns than \texttt{intent}, Table~\ref{app:fullgrid}), so while \emph{some}
classification suffix helps universally, the optimal suffix is a per-probe design choice
rather than a single fixed string. This per-readout tuning is a small residual on top of the
shared-axis effect, not a contradiction of it: the large gain is the no-suffix$\to$suffix
alignment jump (on Gemma, $0.74\to0.93$; App.~\ref{app:mech}), and among \emph{effective}
suffixes the cross-readout differences are ${\lesssim}1$ pt.

\section{Limitations}
\label{sec:limitations}\label{sec:measurement}
\paragraph{Where the effect is weaker.} The classification-\emph{specific} benefit is a
Llama/Gemma result. Qwen is positive vs.\ no-suffix (6/9 datasets, so not a null), but its gain
over a \emph{neutral} suffix is only 4/9 and its precision increment is flat, so we do not claim
the classification-specific effect for it. Per-model readout differs (Llama L31, non-thinking; Qwen L31, thinking; Gemma L47), so
cross-model magnitudes are confounded with readout regime. The content-tailoring result is preliminary (one
genre, source confound), and we do not claim the probe reads ground-truth harmfulness (cf.\
\citealp{Wang2025FalseSecurity}): a content-free format yields a more transferable
\emph{readout}, not a verified concept.

\paragraph{Measurement.} The gain is a ranking improvement, not recalibration: it appears in
threshold-free shared-benign AUC, and accuracy@0.5 understates it (Gemma $+1.8$--$2.8$ vs.\ AUC
$+2.7$--$3.2$); compare prompted detectors on a threshold-free metric, read per-dataset. We prefer
shared-benign AUC over within-dataset AUC, which exists for only the 4 mixed-class benchmarks
(too few to resolve the dissociation; it spuriously reads \texttt{reflect} as $+2.4$ on Qwen);
the shared benign pool recovers all 9 malicious datasets.

\paragraph{Significance.} With $n{=}9$ datasets the minimum two-sided Wilcoxon $p$ is $0.004$, so
per-cell tests have limited power and we do not rest claims on them. We rely instead on the
dataset-level sign-test (22/27 cells positive, $p{<}10^{-3}$), the directional TPR@1\% tests
(one-sided $p{=}0.002$), and the A/B-vs-intent equivalence (TOST, $p{\le}0.005$; Llama and Gemma).

\paragraph{Robustness to layer.} The gain does not depend on the exact readout layer: on Gemma it
is essentially unchanged at a second late layer ($+3.1$ at L40 vs.\ $+3.2$ at L47), and on Qwen it
is near-zero at both L27 and L31, so Qwen's weakness is intrinsic to the model, not the layer.

\section{Conclusion}
\label{sec:conclusion}\label{sec:discussion}
\paragraph{Takeaway.} For prompted activation probes, a classification suffix
consistently improves malicious-input detection; the classification \emph{format}
carries the ranking gain (a content-free criterion suffices), while naming the real
criterion \emph{adds} the low-FPR precision that deployment reads.

\begin{center}\fbox{\begin{minipage}{0.94\linewidth}\small
\textbf{Deploying this.} (1)~append a 29-token post-user classification
suffix via the KV-fork; (2)~train the probe on the end-of-suffix activation as in
standard probing---same cost as a no-suffix probe, no model fine-tuning, no second
prefill (the KV-fork reuses the user prefill; the backbone only supplies the
embedding); (3)~\emph{name the real classes} (malicious/benign),
not placeholders: where the probe has low-FPR headroom (Llama, Gemma) this is
what earns the recall; (4)~evaluate
per-dataset on a threshold-free metric at $N{\ge}2$k; (5)~keep the criterion
\emph{broad}: benchmark-specific content overfits out-of-genre.
\end{minipage}}\end{center}

\paragraph{On ``in the wild.''} This targets agent monitoring: activation probes read the
model's own forward pass on the incoming request to flag malicious inputs before the agent acts
\citep{TaskTracker,PIShield,KramarGeminiProbes2026}, reusing that pass at roughly six orders of
magnitude less compute than a separate LLM monitor \citep{McKenzie2025HighStakes}. We test under
strict LODO, the deployment condition: every number is on a benchmark the probe never saw in
training. One scope caveat: our benchmarks are chat-style, and we do not vary input
\emph{packaging or genre} (single-turn vs.\ multi-turn, tool-call transcripts, or heavily
formatted content such as tables or code), so extension to those inputs is a conjecture,
not a result.

\paragraph{Future work.} The decomposition points to three directions. First,
\textbf{principled suffix optimization}: since a per-readout best rung already beats a fixed
string (\S\ref{sec:arch}) and naive benchmark-specific tailoring overfits out-of-genre, searching
\emph{broad}, transferable attack-family content under out-of-distribution validation is the
natural way to push the gain without overfitting. Second, \textbf{agentic tool-call and
multi-turn inputs}, where the deployment payoff is largest and which our single-turn setup does
not yet test. Third, \textbf{a deeper representational account}: our geometry (App.~\ref{app:mech})
shows \emph{where} the suffix helps but not \emph{why} a content-free classification \emph{format}
reshapes the input's representation to be more separable. Relating this to the format-over-content
effect in in-context learning \citep{Min2022Rethinking} (what asking the model to \emph{classify}
does to its representation of the input, and why the named criterion is largely inert) is an open
question our ablation frames but does not resolve. Because the content-free suffix (\texttt{A/B}) names no criterion, asking the model to \emph{classify} might make the input easier to read along \emph{any} axis it encodes, or it might merely recruit the model's strong safety prior---an effect specific to maliciousness. Reading only the malicious axis we cannot separate these two accounts; applying the same suffix to a non-safety target, such as task type, would.

\section*{Acknowledgements}
We thank the open-source community for the models and datasets this work builds on.

\bibliographystyle{plainnat}
\bibliography{refs}
\appendix
\FloatBarrier
\section{Per-dataset results and cross-model ROC}
\label{app:perds}
All values LODO out-of-fold, $N{=}2000$/dataset. \emph{sb-AUC} = each dataset's malicious examples vs.\ the shared benign pool (absolute; per-fold calibration offsets make deltas, not absolutes, the unit of inference, \S\ref{sec:measurement}).

\begin{table}[h]\centering\caption{\textbf{Llama}: per-dataset shared-benign AUC by suffix.}\small
\begin{tabular}{lcccc}\toprule
dataset & none & reflect & A/B & intent \\ \midrule
deepset & 80.0 & 88.7 & 89.4 & 90.9 \\
harmbench & 80.9 & 88.4 & 90.0 & 91.1 \\
yanismiraoui & 89.2 & 85.9 & 94.6 & 94.0 \\
jayavibhav & 95.4 & 93.8 & 94.3 & 93.5 \\
wildjailbreak & 91.7 & 94.3 & 94.6 & 96.1 \\
qualifire & 92.3 & 95.2 & 97.7 & 95.3 \\
mosscap & 82.8 & 63.4 & 85.4 & 83.9 \\
safeguard & 97.1 & 96.1 & 95.9 & 97.7 \\
advbench & 98.3 & 98.0 & 98.4 & 99.6 \\
\bottomrule\end{tabular}\end{table}
\begin{table}[h]\centering\caption{\textbf{Qwen}: per-dataset shared-benign AUC by suffix.}\small
\begin{tabular}{lcccc}\toprule
dataset & none & reflect & A/B & intent \\ \midrule
deepset & 90.3 & 92.5 & 90.1 & 89.1 \\
harmbench & 89.1 & 88.9 & 94.3 & 92.7 \\
yanismiraoui & 91.3 & 95.7 & 95.4 & 94.1 \\
jayavibhav & 93.7 & 95.0 & 96.3 & 95.6 \\
wildjailbreak & 95.4 & 97.4 & 95.9 & 95.8 \\
qualifire & 98.3 & 98.7 & 98.4 & 97.9 \\
mosscap & 93.2 & 88.9 & 91.4 & 90.2 \\
safeguard & 97.4 & 98.0 & 97.0 & 97.9 \\
advbench & 98.7 & 98.9 & 99.1 & 99.2 \\
\bottomrule\end{tabular}\end{table}
\begin{table}[h]\centering\caption{\textbf{Gemma}: per-dataset shared-benign AUC by suffix.}\small
\begin{tabular}{lcccc}\toprule
dataset & none & reflect & A/B & intent \\ \midrule
deepset & 84.4 & 84.8 & 88.5 & 90.8 \\
harmbench & 85.1 & 84.7 & 93.3 & 92.1 \\
yanismiraoui & 92.5 & 95.4 & 89.8 & 92.4 \\
jayavibhav & 92.6 & 92.8 & 96.1 & 96.4 \\
wildjailbreak & 96.0 & -- & 96.8 & 97.4 \\
qualifire & 96.7 & 95.7 & 97.7 & 97.1 \\
mosscap & 83.2 & 85.0 & 90.4 & 91.6 \\
safeguard & 96.4 & 96.6 & 97.5 & 96.8 \\
advbench & 98.0 & 98.7 & 99.1 & 99.3 \\
\bottomrule\end{tabular}\end{table}

\begin{figure}[H]\centering
\includegraphics[width=0.49\linewidth]{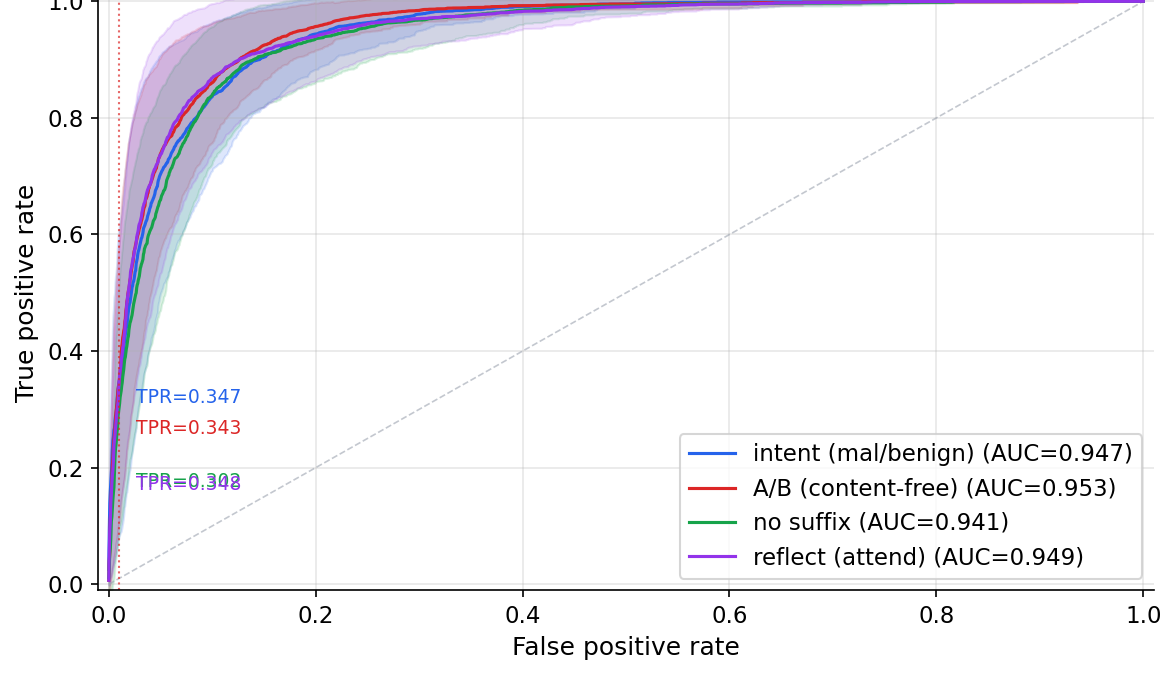}\hfill
\includegraphics[width=0.49\linewidth]{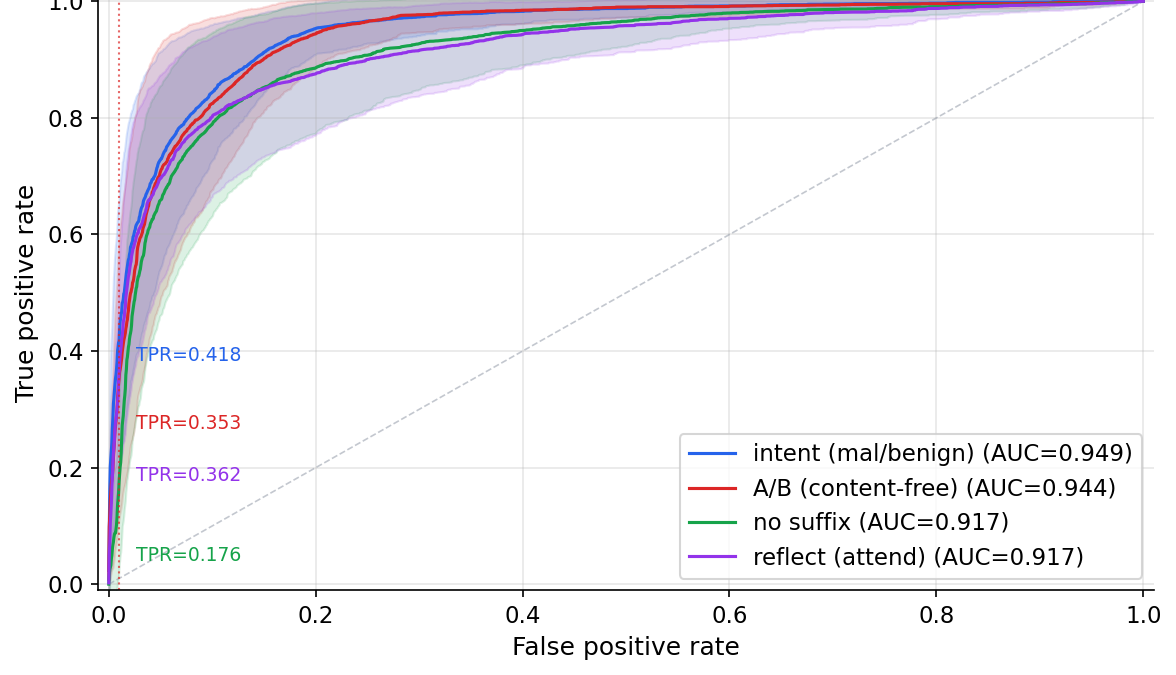}
\caption{\textbf{Qwen (left) and Gemma (right) macro-ROC}, malicious vs.\ shared-benign over the 9 LODO folds (cf.\ Llama, Fig.~\ref{fig:roc}). Qwen curves bunch near a 0.94 baseline; Gemma separates like Llama on area.}\label{fig:roc_appendix}\end{figure}

Across models the pattern is the same: the classification suffixes (\texttt{A/B}, \texttt{intent})
lift the non-saturated datasets most (\texttt{deepset} $80{\to}90$ and \texttt{harmbench} $81{\to}91$ on
Llama) and leave near-ceiling sets (\texttt{advbench}, \texttt{safeguard} ${\approx}98$) unmoved, while
Qwen barely moves anywhere. The macro-ROC curves (Figs.~\ref{fig:roc},~\ref{fig:roc_appendix}) show the
same ranking gain as area under the curve.

\FloatBarrier
\subsection{Precision at a strict operating point (TPR@1\%)}
\label{app:tpr}
At a strict $1\%$-FPR operating point the named criterion earns its keep: \texttt{intent} exceeds
both the content-free \texttt{A/B} suffix and the neutral \texttt{reflect} at TPR@1\% on Llama
($\approx$$+28$pp, 9/9 datasets, one-sided $p{=}0.002$) and Gemma ($+6.6$pp vs.\ A/B, 9/9,
$p{=}0.002$), while Qwen is flat ($+0.3$pp, n.s.). Format alone can lift low-FPR recall (on Gemma
\texttt{reflect} does, $+17$pp), but naming the real classes adds precision on top. TPR@1\% is a
single, high-variance operating point, so we read this increment as directional (consistent 9/9)
rather than a calibrated size, and state the claim as \emph{content adds precision}, not
\emph{content is required} (on Llama, \texttt{A/B} even trends below no-suffix here).

\begin{table}[h]\centering
  \caption{\textbf{Suffix ladder: $\Delta$ TPR@FPR$=$1\% vs.\ \texttt{none}} (pp). \texttt{intent}
    pulls ahead of content-free A/B on Llama/Gemma; A/B \emph{underperforms} no-suffix on Llama;
    Qwen stays flat. The Gemma \texttt{reflect} figure averages 8 datasets (\texttt{wildjailbreak}
    has no \texttt{reflect} run).}
  \label{tab:tpr}\small
  \begin{tabular}{lccc}\toprule
    Suffix & Llama & Qwen & Gemma \\ \midrule
    \texttt{reflect} (attend)     & $-9.1$ & $+4.7$ & $+17.3$ \\
    \texttt{A/B} (content-free)   & $-8.4$ & $+4.2$ & $+17.6$ \\
    \texttt{intent} (mal/benign)  & $\mathbf{+19.6}$ & $+4.5$ & $\mathbf{+24.2}$ \\
    \bottomrule\end{tabular}\end{table}

Per-dataset TPR@1\% (below) shows the increment is broad on Llama and Gemma but dataset-specific
in size; on several Llama datasets the content-free \texttt{A/B} suffix falls below no-suffix,
where \texttt{intent} instead lifts.

\begin{table}[h]\centering\caption{\textbf{Llama}: per-dataset TPR@1\% (\%) by suffix.}\small
\begin{tabular}{lcccc}\toprule
dataset & none & reflect & A/B & intent \\ \midrule
deepset & 5.9 & 10.3 & 5.4 & 30.5 \\
harmbench & 24.2 & 20.8 & 4.5 & 44.8 \\
yanismiraoui & 15.9 & 8.1 & 26.5 & 52.5 \\
jayavibhav & 33.9 & 17.9 & 11.2 & 30.5 \\
wildjailbreak & 18.0 & 14.3 & 10.5 & 42.7 \\
qualifire & 36.4 & 44.2 & 52.9 & 61.2 \\
mosscap & 4.9 & 1.7 & 25.5 & 31.6 \\
safeguard & 61.3 & 40.1 & 29.1 & 66.5 \\
advbench & 77.7 & 39.2 & 36.5 & 94.4 \\
\bottomrule\end{tabular}\end{table}
\begin{table}[h]\centering\caption{\textbf{Qwen}: per-dataset TPR@1\% (\%) by suffix.}\small
\begin{tabular}{lcccc}\toprule
dataset & none & reflect & A/B & intent \\ \midrule
deepset & 8.4 & 8.4 & 9.4 & 4.4 \\
harmbench & 31.0 & 32.2 & 40.8 & 47.2 \\
yanismiraoui & 15.2 & 19.7 & 25.9 & 15.6 \\
jayavibhav & 12.4 & 23.7 & 26.3 & 24.0 \\
wildjailbreak & 12.4 & 30.8 & 13.3 & 19.3 \\
qualifire & 65.4 & 64.7 & 65.7 & 66.7 \\
mosscap & 14.9 & 3.4 & 1.1 & 0.5 \\
safeguard & 51.3 & 65.5 & 54.0 & 57.3 \\
advbench & 60.8 & 65.2 & 72.7 & 77.3 \\
\bottomrule\end{tabular}\end{table}
\begin{table}[h]\centering\caption{\textbf{Gemma}: per-dataset TPR@1\% (\%) by suffix.}\small
\begin{tabular}{lcccc}\toprule
dataset & none & reflect & A/B & intent \\ \midrule
deepset & 11.8 & 19.2 & 14.8 & 21.7 \\
harmbench & 16.8 & 37.2 & 46.8 & 62.0 \\
yanismiraoui & 17.9 & 49.4 & 23.0 & 27.5 \\
jayavibhav & 5.2 & 15.3 & 13.3 & 26.8 \\
wildjailbreak & 7.8 & -- & 32.0 & 35.3 \\
qualifire & 54.1 & 54.8 & 63.4 & 68.7 \\
mosscap & 2.2 & 13.6 & 1.1 & 3.8 \\
safeguard & 23.1 & 39.6 & 43.7 & 48.2 \\
advbench & 21.0 & 61.5 & 80.0 & 83.7 \\
\bottomrule\end{tabular}\end{table}

\FloatBarrier
\section{Benchmarks}
\label{app:data}
The 13 evaluation benchmarks, with role and sizes in the primary panel (up to $N{=}2000$/dataset; counts vary slightly by model). Nine contain malicious examples (scored against the shared benign pool); the four benign-source sets contribute to that pool.
\begin{table}[h]\centering\small
\caption{\textbf{Evaluation benchmarks.} Role and number of benign/malicious examples per benchmark in the primary panel.}\label{tab:benchmarks}
\begin{tabular}{llrr}\toprule
Benchmark & Type & benign & malicious \\ \midrule
advbench & harmful instructions & --- & 520 \\
harmbench & harmful behaviors & --- & 400 \\
wildjailbreak & jailbreaks & --- & 2000 \\
yanismiraoui & jailbreaks & --- & 1034 \\
deepset & prompt injection & 343 & 203 \\
jayavibhav & prompt injection & 996 & 1004 \\
mosscap & prompt injection/leak & --- & 2000 \\
qualifire & safety (mixed) & 1205 & 795 \\
safeguard & safety (mixed) & 1380 & 620 \\
10k\_prompts\_ranked & benign chat & 2000 & --- \\
dolly\_15k & benign instructions & 2000 & --- \\
openorca & benign chat & 2000 & --- \\
softAge & benign & 1001 & --- \\
\bottomrule\end{tabular}\end{table}

\FloatBarrier
\section{Probe architectures and settings}
\label{app:probes}
All probes are trained per LODO fold on frozen, standardized residual-stream activations at the readout layer. The single-position probe is L2-logistic ($C{=}0.1$) at the last suffix token; the multi-position pooling probes follow the production ``Gemini probe'' family \citep{KramarGeminiProbes2026} (over all token positions, Adam). Readout layers: Llama L31, Qwen L27/L31, Gemma L40/L47.
\begin{table}[h]\centering\small
\caption{\textbf{Probe readouts.} Token positions each readout uses and how it is fit.}\label{tab:probes}
\begin{tabular}{lll}\toprule
Readout & Positions & Fit \\ \midrule
single-position (linear) & last suffix token & L2-logistic, $C{=}0.1$ \\
attention pooling & all & softmax-weighted, Adam \\
multi-max pooling & all & top-$k$ max, Adam \\
MLP pooling & all & per-position MLP + pool, Adam \\
\bottomrule\end{tabular}\end{table}

\FloatBarrier
\subsection{Full architecture $\times$ suffix grid}
\label{app:fullgrid-sec}
Table~\ref{app:fullgrid} reports every cell behind the architecture-agnostic result (\S\ref{sec:arch}):
six suffix conditions $\times$ four pooling readouts $\times$ three families $\times$ two layers. A
classification suffix lifts almost every column over \texttt{none} where the baseline has headroom;
the best-performing rung is readout-dependent (\texttt{length}/\texttt{reflect} top more columns than
\texttt{intent}), and Qwen is near-ceiling throughout.

\begin{table}[h]\centering
\caption{\textbf{Full architecture $\times$ suffix grid.} Shared-benign LODO AUC (\%), $N{=}2000$/dataset, 9 malicious datasets vs.\ shared benign pool. Each block is one model/layer; rows are the suffix ladder, columns the pooling readouts; \texttt{none} is the matched baseline, bold = best rung per column. Almost every column beats \texttt{none} where the baseline has headroom; the winning rung is readout-dependent.}
\label{app:fullgrid}\small
\begin{tabular}{lcccc}
\toprule
\multicolumn{5}{l}{\textbf{Gemma-4 (L47)}}\\
Readout & attention & multi-max & MLP & mean\\ \midrule
none & 93.6 & 90.8 & 93.1 & 90.9\\
irrelevant & 94.8 & 95.3 & 94.0 & 92.2\\
reflect & 95.1 & \textbf{95.5} & 94.2 & 92.9\\
classify:length & \textbf{95.6} & 95.0 & \textbf{95.0} & 93.2\\
classify:A/B & 95.2 & 95.0 & 94.9 & 93.2\\
classify:intent & 95.6 & 95.4 & 95.0 & \textbf{93.3}\\
\midrule
\multicolumn{5}{l}{\textbf{Gemma-4 (L40)}}\\
Readout & attention & multi-max & MLP & mean\\ \midrule
none & 95.8 & 95.6 & 94.8 & 93.9\\
irrelevant & 95.3 & 94.5 & 95.2 & 94.5\\
reflect & 95.6 & 95.0 & 95.5 & \textbf{95.2}\\
classify:length & 96.3 & 95.7 & \textbf{95.9} & 94.8\\
classify:A/B & \textbf{96.8} & 95.5 & 95.7 & 94.8\\
classify:intent & 96.7 & \textbf{97.0} & 95.8 & 94.8\\
\midrule
\multicolumn{5}{l}{\textbf{Qwen3.5 (L31)}}\\
Readout & attention & multi-max & MLP & mean\\ \midrule
none & 94.6 & 95.0 & 95.1 & 94.7\\
irrelevant & \textbf{96.0} & \textbf{95.7} & 96.0 & 95.2\\
reflect & 95.7 & 95.4 & 95.9 & 95.2\\
classify:length & 95.3 & 95.3 & \textbf{96.1} & \textbf{95.4}\\
classify:A/B & 94.9 & 95.1 & 96.0 & 95.3\\
classify:intent & 95.3 & 95.2 & 95.8 & 95.2\\
\midrule
\multicolumn{5}{l}{\textbf{Qwen3.5 (L27)}}\\
Readout & attention & multi-max & MLP & mean\\ \midrule
none & \textbf{96.8} & 94.9 & 95.4 & 95.2\\
irrelevant & 95.4 & 95.2 & \textbf{96.3} & 95.9\\
reflect & 95.3 & 95.4 & 96.2 & \textbf{96.0}\\
classify:length & 96.6 & \textbf{96.3} & 96.2 & 95.9\\
classify:A/B & 95.8 & 95.3 & 96.2 & 95.9\\
classify:intent & 94.9 & 96.2 & 96.0 & 95.8\\
\midrule
\multicolumn{5}{l}{\textbf{Llama-3.1 (L31)}}\\
Readout & attention & multi-max & MLP & mean\\ \midrule
none & 92.8 & 92.9 & 93.4 & 92.0\\
irrelevant & 94.4 & 94.7 & 94.5 & 93.5\\
reflect & 94.5 & 95.5 & 95.1 & \textbf{94.2}\\
classify:length & 95.5 & \textbf{95.8} & 95.2 & 94.0\\
classify:A/B & \textbf{95.6} & 94.4 & \textbf{95.2} & 94.0\\
classify:intent & 95.3 & 94.2 & 95.2 & 93.8\\
\midrule
\multicolumn{5}{l}{\textbf{Llama-3.1 (L27)}}\\
Readout & attention & multi-max & MLP & mean\\ \midrule
none & 93.3 & 94.1 & 94.8 & 93.5\\
irrelevant & \textbf{95.8} & 94.7 & 95.5 & 94.5\\
reflect & 95.8 & \textbf{95.2} & \textbf{95.9} & \textbf{95.0}\\
classify:length & 95.5 & 94.8 & 95.8 & 94.8\\
classify:A/B & 95.5 & 95.2 & 95.8 & 94.7\\
classify:intent & 94.9 & 95.2 & 95.8 & 94.6\\
\bottomrule
\end{tabular}\end{table}

\FloatBarrier
\subsection{What the suffix does to the representation (where there is headroom)}
\label{app:mech}

\begin{figure}[H]
  \centering
  \includegraphics[width=0.9\linewidth]{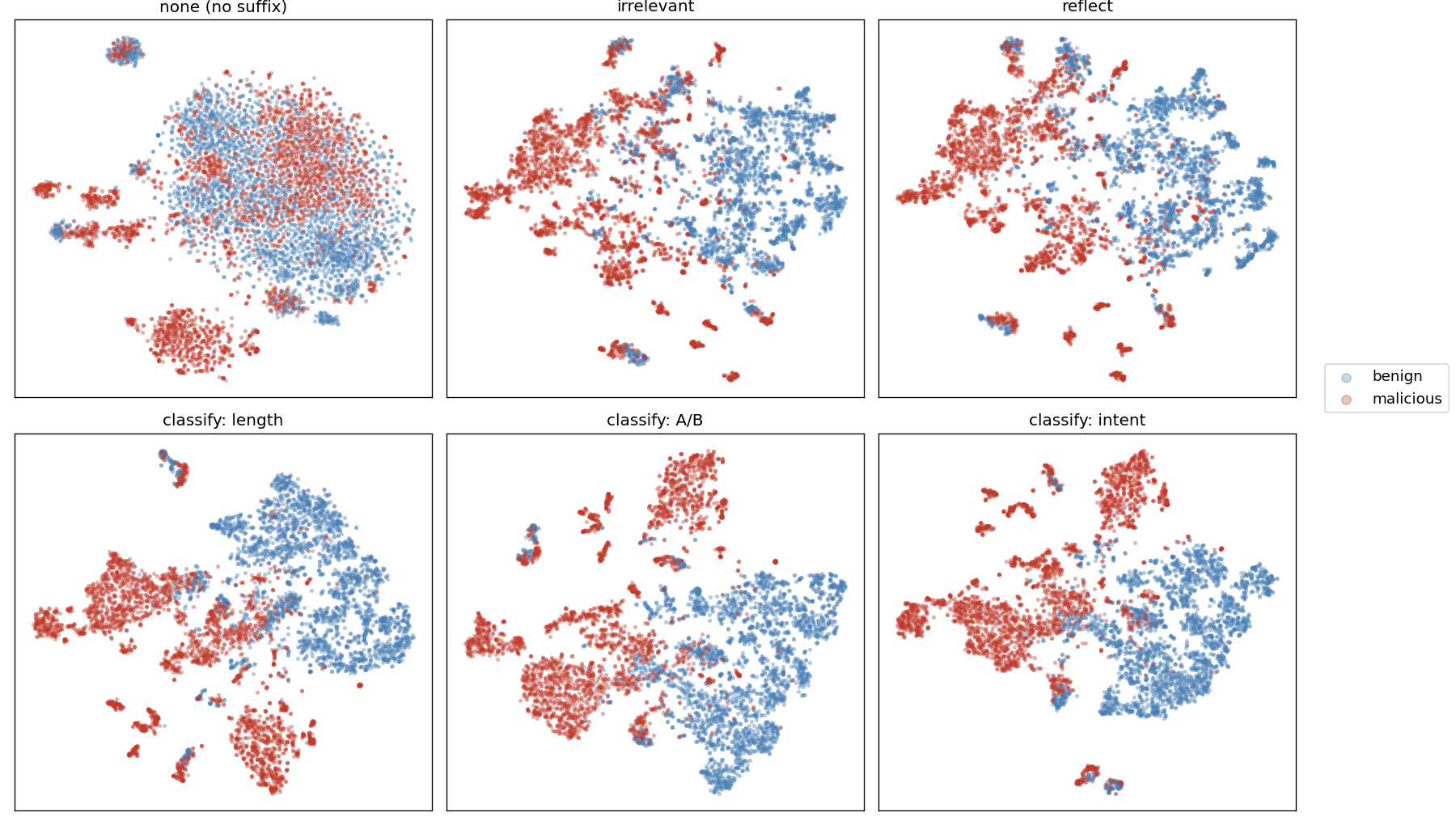}\\[6pt]
  \includegraphics[width=0.55\linewidth]{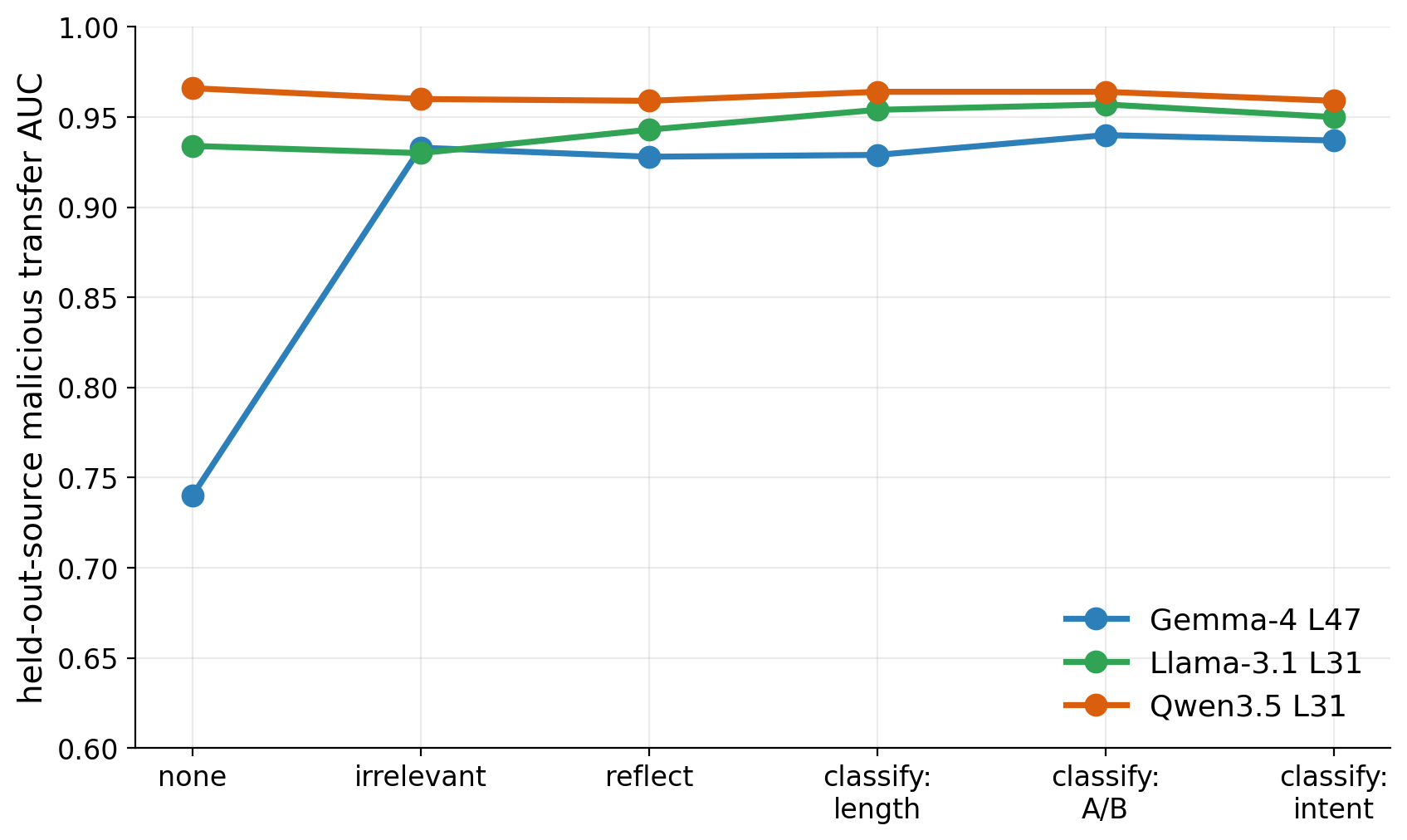}
  \caption{\textbf{Where it has headroom, the suffix aligns the malicious representation (Gemma).}
    \emph{Top:} t-SNE of Gemma's single-position readout across the ladder, colored
    malicious/benign; classes interleave with no suffix and separate under the classification
    suffixes. \emph{Bottom:} leave-one-source-out transfer of a mean-difference malicious axis.
    Gemma climbs $0.74\to0.93$; Llama and Qwen start already-aligned ($0.93$, $0.96$) and barely
    move.}
  \label{fig:mech}
\end{figure}

This appendix expands the one-sentence mechanism in \S\ref{sec:ladder}. We look two ways at the
single-position readout, both independent of the trained probe. \textbf{(i) Cross-source axis.}
We form a malicious direction as the mean difference (malicious $-$ benign) over all-but-one
attack source and test whether it separates the held-out source's malicious examples from the
shared benign pool: a leave-one-\emph{source}-out transfer of the \emph{axis itself}, not of a
fitted probe. On Gemma this transfer AUC climbs from $0.74$ with no suffix to ${\sim}0.93$ once
\emph{any} suffix is appended (Fig.~\ref{fig:mech}, right): without a suffix the per-source
malicious directions are nearly orthogonal (mean pairwise cosine $0.18$), and the suffix rotates
them into near-alignment ($0.90$ for the classification rungs). \textbf{(ii) A direct look.} A
t-SNE of Gemma's readout colored by label (Fig.~\ref{fig:mech}, left) shows the same thing: with
no suffix, malicious and benign interleave; under the classification suffix they separate.

This is a \emph{headroom} effect, not a universal transformation. Llama and Qwen already carry a
well-aligned malicious axis with no suffix (transfer AUC $0.93$ and $0.96$; Fig.~\ref{fig:mech},
bottom), so there is little left for a suffix to build, precisely why their detection gains are the smallest of the three
(\S\ref{sec:arch}). The t-SNE is illustrative; the transfer AUC is the quantitative claim.

\FloatBarrier
\subsection{Format/content dissociation and ROC detail}
\label{app:dissoc}

\begin{figure}[H]
  \centering
  \includegraphics[width=\linewidth]{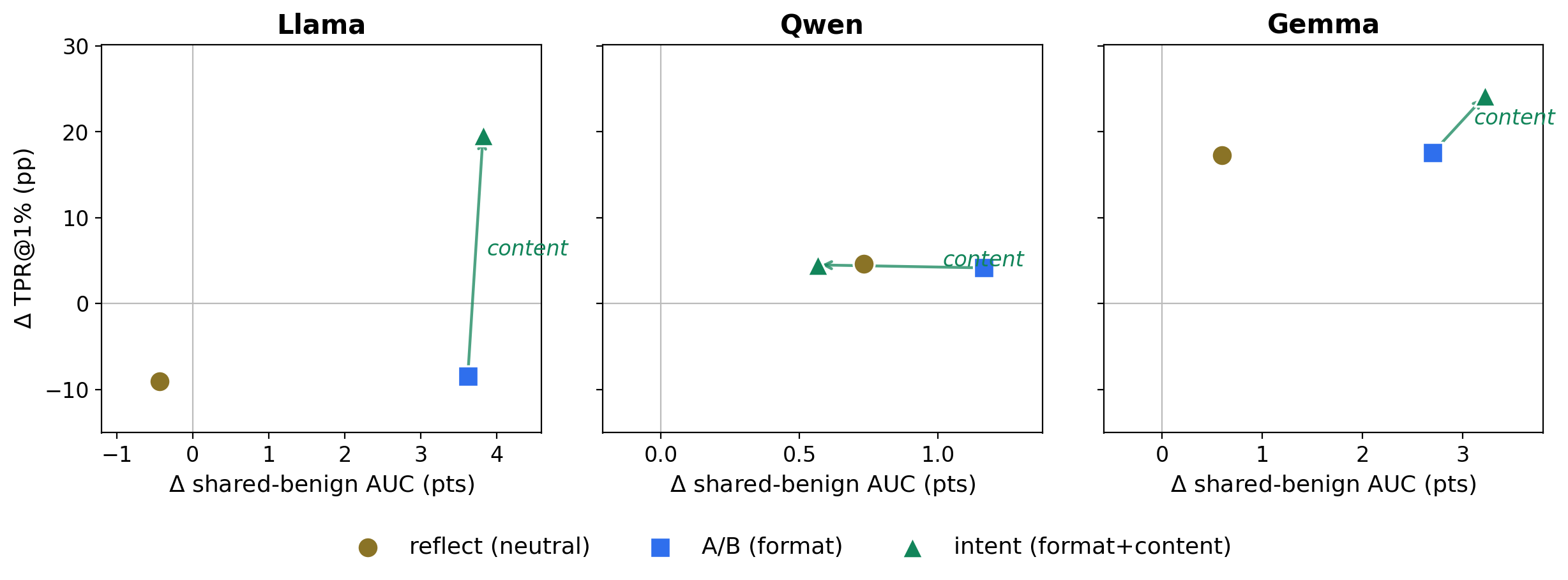}
  \caption{\textbf{The format/content dissociation.} Each point is a suffix, by its mean change
    vs.\ no-suffix in shared-benign AUC ($x$, ranking) and TPR@1\% ($y$, precision); one panel per
    model. \texttt{A/B} and \texttt{intent} share the same $x$ but \texttt{intent} is far higher on
    $y$ ($\approx$$28$pp on Llama); neutral \texttt{reflect} sits near the origin (except Gemma,
    where it already lifts precision).}
  \label{fig:dissoc}
\end{figure}

\begin{figure}[H]
  \centering
  \includegraphics[width=0.6\linewidth]{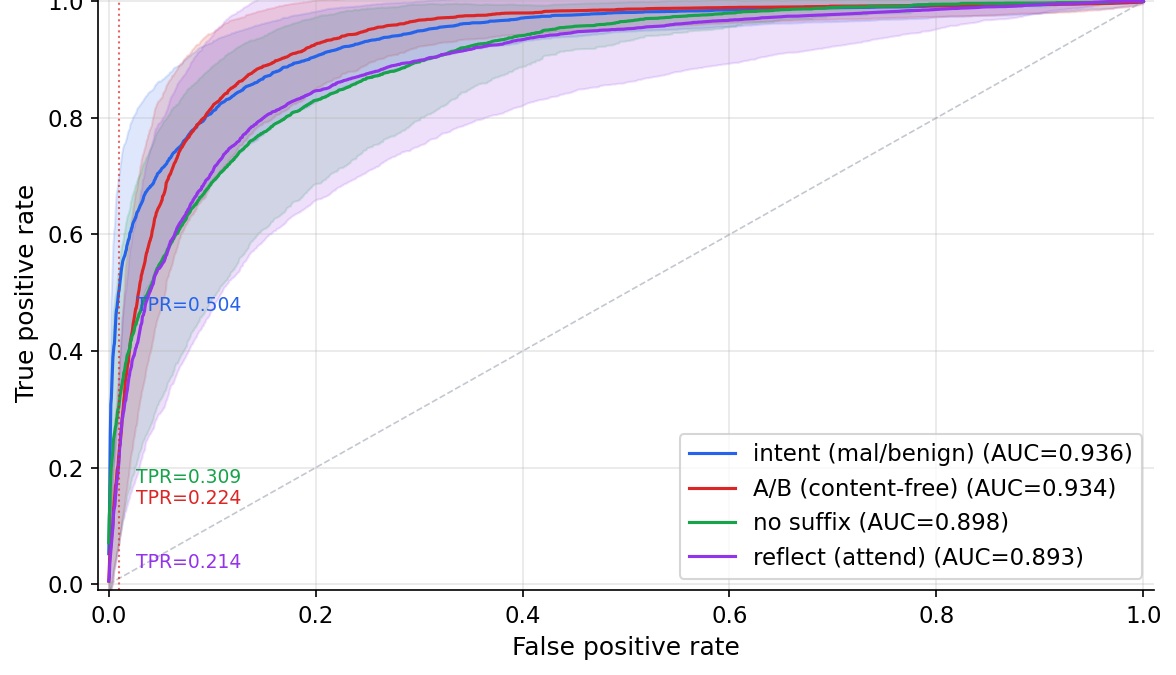}
  \caption{\textbf{Llama macro-ROC (malicious vs.\ shared benign, 9 LODO folds).} \texttt{intent}
    (0.936) and \texttt{A/B} (0.934) lift over no-suffix (0.898); \texttt{reflect} (0.893) does not.
    Curves cross near FPR$=$1\% (dotted): \texttt{A/B} matches \texttt{intent} on area but drops
    below no-suffix at the strict operating point. $\pm1\sigma$ $=$ cross-dataset spread.}
  \label{fig:roc}
\end{figure}

Figure~\ref{fig:dissoc} plots each suffix by its mean change vs.\ no-suffix in shared-benign
AUC (ranking) and TPR@1\% (precision), one panel per model: the content-free \texttt{A/B} and
named \texttt{intent} suffixes share the same ranking gain ($x$) but \texttt{intent} is far
higher on precision ($y$): the format/content dissociation. Figure~\ref{fig:roc} shows the
Llama macro-ROC: \texttt{A/B} matches \texttt{intent} on area yet drops below no-suffix at the
strict $1\%$-FPR operating point, where naming the real criterion helps most.

\FloatBarrier

\end{document}